\documentclass[letterpaper, 10 pt, conference]{ieeeconf}  

\IEEEoverridecommandlockouts                              

\usepackage{amsmath}
\usepackage{cite}
\usepackage{graphicx}
\usepackage{algorithm} 				
\usepackage{algpseudocode}			

\usepackage{float}  
\usepackage{subfigure}  
\usepackage{booktabs}
\usepackage{threeparttable} 
\usepackage{adjustbox} 
\usepackage{soul}
\usepackage{multirow}

\title{\LARGE \bf
Accelerated Quasi-Static FEM for Real-Time Modeling of Continuum Robots with Multiple Contacts and Large Deformation
}

\author{Hao Chen, Jian Chen, Xinran Liu, Zihui Zhang, Yuanrui Huang, Zhongkai Zhang, and Hongbin Liu
\thanks{*This work was supported by the Institute of Automation, Chinese Academy of Sciences and the InnoHK programme. (Corresponding author: Hongbin Liu)}
\thanks{Hao Chen, Xinran Liu, and Yuanrui Huang are with the School of Artificial Intelligence, University of Chinese Academy of Sciences, Beijing 100049, China. (e-mail: chen.hao2020@ia.ac.cn; liuxinran2020@ia.ac.cn; huangyuanrui2020@ia.ac.cn)}%
\thanks{Zihui Zhang is with the Institute of Automation, Chinese Academy of Sciences, Beijing 100190, China. (e-mail: zihui.zhang@ia.ac.cn)}%
\thanks{Zhongkai Zhang, Jian Chen, and Hongbin Liu are with the Centre of AI and Robotics (CAIR), Hong Kong Institute of Science Innovation, Chinese Academy of Sciences, Hongkong, China. (e-mail: zhongkai.zhangzkz@gmail.com; jian.chen@cair-cas.org.hk; liuhongbin@ia.ac.cn)}%
}

\begin{document}

\maketitle
\thispagestyle{empty}
\pagestyle{empty}

\begin{abstract}

Continuum robots offer high flexibility and multiple degrees of freedom, making them ideal for navigating narrow lumens. However, accurately modeling their behavior under large deformations and frequent environmental contacts remains challenging. Current methods for solving the deformation of these robots, such as the  Model Order Reduction and Gauss-Seidel (GS) methods, suffer from significant drawbacks. They experience reduced computational speed as the number of contact points increases and struggle to balance speed with model accuracy. To overcome these limitations, we introduce a novel finite element method (FEM) named Acc-FEM. Acc-FEM employs a large deformation quasi-static finite element model and integrates an accelerated solver scheme to handle multi-contact simulations efficiently. Additionally, it utilizes parallel computing with Graphics Processing Units (GPU) for real-time updates of the finite element models and collision detection. Extensive numerical experiments demonstrate that Acc-FEM significantly improves computational efficiency in modeling continuum robots with multiple contacts while achieving satisfactory accuracy, addressing the deficiencies of existing methods.

\end{abstract}


\section{INTRODUCTION}



Continuum robots (CR) have gained significant attention in robotics due to their unique advantages, such as operational safety and environmental adaptability \cite{ref01,ref02,ref03}. However, CR research faces several challenges, particularly in narrow human body cavities where the robot frequently contacts the cavity wall. Surgeons require accurate information on the stress and deformation of CR during surgery. While shape-sensing techniques have emerged as a potential solution, sensors like Fiber Bragg Gratings and electromagnetic tracking techniques are limited by size, cost, or complex assembly requirements. Additionally, intraoperative imaging modalities, such as fluoroscopy, endoscopy, and ultrasound, often require bulky sensing equipment \cite{ref08}. Real-time detection of all contact forces and deformations of CR remains a significant challenge \cite{ref09}.

     \begin{figure}[t]
            \vspace{2mm}
		\centering
		\includegraphics[width=0.48\textwidth]{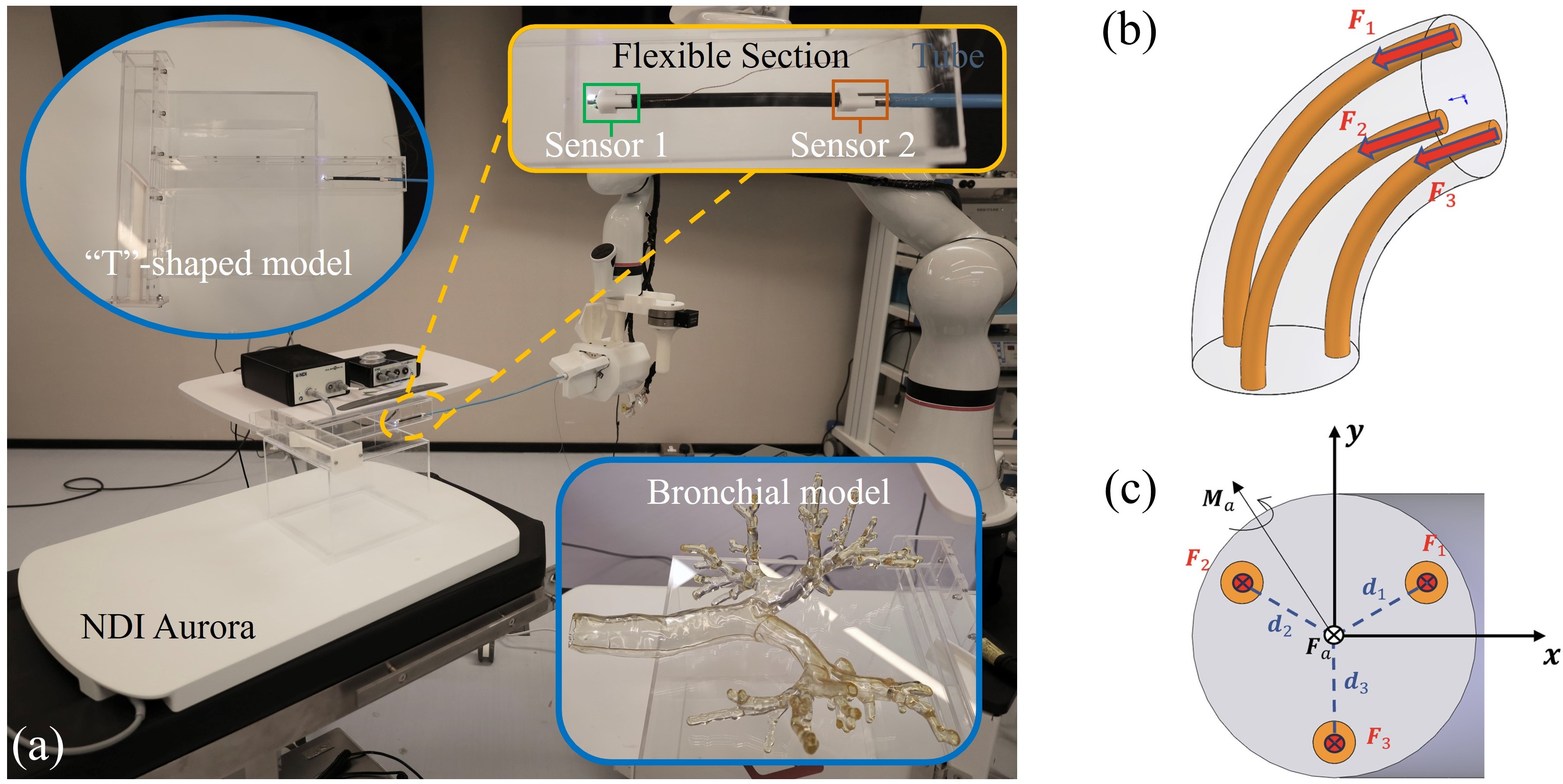}
		\caption{Continuum robot for bronchial surgery. (a) The continuum robot, mounted on a robotic arm, has two position sensors at its end. (b) At the end of the robot, three cables are connected to drive the robot's steering. (c) The tension $\boldsymbol{F}_{i}$ of the cables can be translated to the center of the section into a force $\boldsymbol{F}_{a}$ and a moment $\boldsymbol{M}_{a}$.}
		\label{figurelabel1}
		\vspace{-2mm}
	\end{figure}

Physical simulation-based contact force estimation is a promising technique for continuum robots. However, accurately simulating the deformation and contact forces between the continuum robot and its surroundings in real-time remains a challenge due to the robot's infinite degrees of freedom and exposure to multiple actuations and contact forces. The finite element method is widely used in modeling continuum robots and medical devices, but analyzing a system with an infinite-dimensional full-order model is difficult\cite{ref14}.

To achieve real-time computation of FEM with multiple contact constraints, various methods have been employed. One approach projects the motion of continuum robots from their task space into a contact space, where the FEM is iteratively solved using GS methods in a lower-dimensional space\cite{ref15}. Finite element software frameworks like SOFA\cite{ref16,ref17,ref18,ref19,ref20,ref21} and physics engines like MuJoCo have been developed to support the modeling, simulation, and control of soft robots. However, computational speed decreases as the number of contacts and actuators increases. Another approach uses the Model Order Reduction technique to map the dynamic model of a soft robot to a reduced model of lower dimension\cite{ref22}. Some researchers aim to simplify contact interactions by implementing bounding-volume hierarchies and axis-aligned bounding boxes to streamline collision detection\cite{ref23,ref24}, but this may not completely prevent penetration\cite{ref25}.

These methods, which prioritize fast computation over accuracy, are not effectively addressing the challenge of FEM calculation speed when dealing with multiple contact constraints. The primary motivation of this paper is to investigate a novel solution that can maintain fast computation speed regardless of the increasing number of contacts or the dimension of the FEM. This paper proposes the Acc-FEM, an accelerated quasi-static FEM for modeling a continuum robot with multiple contacts and large deformations, which takes into account external forces such as actuation and contact forces. The continuum robot is approximated using multiple beam elements directly connected through global nodes and its displacement is represented through interpolation of global node displacement, while external forces are represented as node forces. The FEM problem is formulated based on Newton's first law. To effectively address the real-time computing challenges of simulating a continuum robot, an accelerated solver scheme is presented. Additionally, A parallel computing method based on GPU is proposed for real-time updating of finite element models and collision detection algorithms. This accelerated approach enables accurate and efficient modeling of a continuum robot's behavior under multi-contact conditions, which has potential for use in a wide range of applications.

The rest of this paper is organized as follows. Section II details the formulation of the continuum robot's motion with contact. Section III presents a fast solver scheme for solving the simulation problem. In Section IV, several numerical examples are provided to evaluate the model. Section V summarizes the proposed methods and experimental results. Finally, Section VI concludes the paper.

\section{MODEL FORMULATION}

In this section, the continuum robot is introduced, and the FEM of the continuum robot and the formulation of the force from contact positions, known fixed nodes, and actuations are presented.

\subsection{The Continuum Robot}

The FEM method proposed in this paper is specifically designed for cable-driven continuum robots used in minimally invasive surgery. The continuum robot discussed here is an adaptation of the model described in \cite{ref26}, but it is configured to include only the inner endoscope. As shown in Fig.\ref{figurelabel1}(a), the robot's body is a long tube made of rubber that deforms when subjected to force. At the end of the flexible robot, a camera and a surgical tool are attached for observation and operation. As illustrated in Fig.\ref{figurelabel1}(b) and (c), the robot features a 70mm long hyoid bone, which is significantly less rigid compared to the rest of the structure. Three nitinol cables, controlled by brushless coreless motors (ASSUN), are welded to the distal end of the robot, facilitating its flexible movement. Additionally, a stepper motor is employed to facilitate the robot's insertion.

    

\subsection{FEM Model of Continuum Robot}

The robot\textquotesingle s centerline is substantially greater than its cross-sectional area. Therefore, we use beam elements to discretize the robot, with nodes distributed along the center line and each node having six degrees of freedom. We divide the robot into $M$ beam units of equal section length \(l_{0}\), resulting in a total length of \(M \times l_{0}\). For this paper, we consider the robot to have low speed and thus model it using quasi-static equations.

For the continuum robot, the quasi-static equation of motion can be derived from Newton's first law:
\begin{equation}
\label{eqn01}
\boldsymbol{F}\left(\boldsymbol{x}_t\right)-\boldsymbol{F}_{ext}\left(\boldsymbol{x}_t\right)=\boldsymbol{0}
\end{equation}
where \(\boldsymbol{F}\) and \(\boldsymbol{F}_{ext}\) are the assembled vectors of nodal internal and external force respectively. \(\boldsymbol{F}_{ext}\) represents the external force. \(\boldsymbol{x}_{t}\) is the vector of position, whose subscript represents the time step.

\subsection{Formulation of Contact Force}

Firstly, we introduce the definition of \(\boldsymbol{F}_{ext}\), which can be expressed as the sum of external forces:
\begin{equation}
\label{eqn02}
\boldsymbol{F}_{ext} = \boldsymbol{F}_{c}+\boldsymbol{F}_{fn}+\boldsymbol{F}_{a}
\end{equation}
where \(\boldsymbol{F}_{c}\), \(\boldsymbol{F}_{fn}\) and \(\boldsymbol{F}_{a}\) are the contact force, the force acting on the known fixed nodes and the force of actuation, respectively.

In simulation, the 3D mesh of the human natural cavity, derived from preoperative MRI reconstruction, is composed of multiple triangles. Let the normal vector of a contact plane triangle be represented by \(\boldsymbol{n}_{i}\), where \(i\) represents the \(i\)-th contact point. Assume that the direction of \(\boldsymbol{n}_{\boldsymbol{i}}\) has been adjusted to point towards the half-space which is divided by the contact plane on the other side of the robot. A point on the contact plane is denoted as \(\boldsymbol{X}_{I}\), and the contact is given by
\begin{equation}
\label{eqn03}
\boldsymbol{n}_{i}^{T}\boldsymbol{N}_{i}(\xi)\boldsymbol{B}_{i}\left( \boldsymbol{x}_{t} + \boldsymbol{S}_{0} \right) \leq \boldsymbol{n}_{i}^{T}\boldsymbol{X}_{I}
\end{equation}
where \(\boldsymbol{S}_{0}\) denotes the global position of the node of the continuum robot in its initial state. And \(\boldsymbol{N}_{i}(\xi)\) is the FEM interpolation function of the contact point\cite{ref27}, considering the continuity requirements and computational cost of the beam element, which is defined as:
\begin{equation}
\label{eqn04}
\boldsymbol{N}_{i}(\xi)=\begin{bmatrix}
P_{1} & 0 & 0 & P_{2} & 0 & 0 \\
0 & P_{3} & 0 & 0 & P_{4} & 0 \\
0 & 0 & P_{3} & 0 & 0 & P_{4}
\end{bmatrix}
\end{equation}
where \(\xi \in \left\lbrack 0,l_{0} \right\rbrack\) is the distance from the contact point to the node on the element closer to the base and \(P_{k},k = 1,\cdots,4\) is the shape function, given by
\begin{gather}
\label{eqn05}
P_{1}(\xi) = 1 - \frac{\xi}{l_{0}},\ P_{2}(\xi) = \frac{\xi}{l_{0}} \\
\label{eqn06}
P_{3}(\xi) = 1 - \frac{3\xi^{2}}{l_{0}^{2}} + \frac{2\xi^{3}}{l_{0}^{3}},\ 
P_{4}(\xi) = \frac{3\xi^{2}}{l_{0}^{2}} - \frac{2\xi^{3}}{l_{0}^{3}}
\end{gather}
and \(\boldsymbol{B}_{i}\) is a matrix that extracts the element containing the contact point from the global configuration, which can be represented as

\begin{equation}
\label{eqn07}
\boldsymbol{B}_{i} =\begin{bmatrix} \boldsymbol{0}_{3 \times m_1}  & \boldsymbol{I}_{3 \times 3} & \boldsymbol{0}_{3 \times 3} & \boldsymbol{0}_{3 \times 3} & \boldsymbol{0}_{3 \times m_2}\\ \boldsymbol{0}_{3 \times m_1}  & \boldsymbol{0}_{3 \times 3} & \boldsymbol{0}_{3 \times 3} & \boldsymbol{I}_{3 \times 3} & \boldsymbol{0}_{3 \times m_2}
\end{bmatrix}
\end{equation}
where $m_1 = 6m - 6$ and $m_2 = 6M - 6m - 3$. And \(M\) is the node number of the FEM model and \(m\) represents the serial number of the element with the contact point. \(\boldsymbol{I}_{3 \times 3}\) is the identity matrix and \(\boldsymbol{0}_{3 \times 3}\) is the zero matrix. 

Combining all contact constraints (\ref{eqn03}) yields (\ref{eqn08}) with (\ref{eqn09}) and (\ref{eqn10}), which is
\begin{gather}
\label{eqn08}
\boldsymbol{A}_{c}\Delta\boldsymbol{x}_{t} - \boldsymbol{b}_{c} \leq \boldsymbol{0} \\
\label{eqn09}
\boldsymbol{A}_{c} = \sum_{i}^{}{\boldsymbol{e}_{i}\boldsymbol{n}_{i}^{T}\boldsymbol{N}_{i}\boldsymbol{B}_{i}} \\
\label{eqn10}
\small
\boldsymbol{b}_{c}=\sum_{i}^{}{\boldsymbol{e}_{i}\left( \boldsymbol{n}_{i}^{T}\boldsymbol{X}_{I} - \boldsymbol{n}_{i}^{T}\boldsymbol{N}_{i}\boldsymbol{B}_{i}\boldsymbol{S}_{0} - \boldsymbol{n}_{i}^{T}\boldsymbol{N}_{i}\boldsymbol{B}_{i}\boldsymbol{x}_{t - 1} \right)}
\end{gather}
where \(\boldsymbol{e}_{i}\) is the \(i\)-th column vector of the identity matrix \(\boldsymbol{I}_{3 \times 3}\). And \(\Delta\boldsymbol{x}_{t}=\boldsymbol{x}_{t}-\boldsymbol{x}_{t - 1}\) is the global nodal displacement between time \(t - 1\) and time \(t\), with six degrees of freedom at each node. \(\boldsymbol{A}_{c}\) is a sparse matrix which gathers contact directions, and \(\boldsymbol{b}_{c}\) represents the distance between robot and contact triangles. Let the global contact force vector \(\boldsymbol{F}_{c}\) be:
\begin{equation}
\label{eqn11}
\boldsymbol{F}_{c} = - \boldsymbol{A}_{c}^{T}\boldsymbol{y}_{c} = - \sum_{i}^{}{\boldsymbol{B}_{i}^{T}\boldsymbol{N}_{i}^{T}\boldsymbol{n}_{i}\boldsymbol{e}_{i}^{T}\boldsymbol{y}_{c}}
\end{equation}
where \(\boldsymbol{y}_{c}\) represents the contact force induced by each contact, and serves as the Lagrange multiplier corresponding to the constraint, whose component can be zero. Derived from the Signorini Conditions\cite{ref14},\cite{ref18} could get
\begin{equation}
\label{eqn12}
\boldsymbol{0} \leq \boldsymbol{y}_{c}\bot(\boldsymbol{A}_{c}\Delta\boldsymbol{x}_{t} - \boldsymbol{b}_{c}) \leq \boldsymbol{0}
\end{equation}
where the symbol \(\bot\) signifies
\begin{equation}
\label{eqn13}
\boldsymbol{Diag}\left( \boldsymbol{y}_{c} \right)\left( \boldsymbol{A}_{c}\boldsymbol{x}_{t} - \boldsymbol{b}_{c} \right) = \boldsymbol{0}
\end{equation}

\subsection{Fixed Nodes Constraints}\label{d.-fixed-nodes-constraints}

Similar to the contact case, constraints are defined with Lagrange multipliers on the boundary conditions to represent the fixed nodes of the continuum surgical robot. The equation representing the known positions where the continuum robot is fixed is defined as an equality constraint:
\begin{equation}
\label{eqn14}
\boldsymbol{y}_{fn}\bot\left( \boldsymbol{A}_{fn}\Delta\boldsymbol{x}_{t} - \boldsymbol{b}_{fn} \right) = \boldsymbol{0}
\end{equation}
where \(\boldsymbol{y}_{fn}\) is the Lagrange multiplier corresponding to the equality constraints. It is important to note that the component of \(\boldsymbol{y}_{fn}\) can be arbitrary. \(\boldsymbol{A}_{fn}\) is a sparse matrix gathering possible directions of the force that holds the fixed position. Therefore, the force acting on the fixed positions is defined as:
\begin{equation}
\label{eqn15}
\boldsymbol{F}_{fn} = - \boldsymbol{A}_{fn}^{T}\boldsymbol{y}_{fn}
\end{equation}

\subsection{Actuation Constraints}\label{-actuation-constraints}

Since the cable of the robot is fixed at the end of the robot's hyoid bone, as shown in Fig.\ref{figurelabel1}(b), the deformation of the robot when subjected to cable tension also mainly occurs here. Therefore, when considering the actuation of the robot, we assume that all forces and moments are located at the end of the robot, that is, at the last node of the last beam unit. The fixing method of the end cable is shown in Fig.\ref{figurelabel1}(c).

Initially, the robot is segmented into beam elements and assumes that the nodes are distributed along the centerline of the robot. \(\boldsymbol{M}_{j}\), where $j$ ranges from 1 to 3,  is generated by the translation of the tension in the $j$-th cable. Using the theorem of force translation\cite{ref28}, could get:
\begin{equation}
\label{eqn16}
\boldsymbol{M}_{j} = \boldsymbol{d}_{j}\times\boldsymbol{F}_{j}
\end{equation}
with \(\boldsymbol{d}_{j}\) a vector from the node to the cable connection point. It is worth noting that \(\boldsymbol{d}_{j}\), \(\boldsymbol{F}_{j}\) and \(\boldsymbol{M}_{j}\) are \(3 \times 1\) vectors and are defined in the local coordinate system. Then the tension of the cables is translated to the nodes. \(\boldsymbol{F}_{a}\) is expressed as:
\begin{equation}
\label{eqn17}
\boldsymbol{F}_{a} = \sum_{j = 1}^{3}{(\boldsymbol{R}_{M,1}^{T}\boldsymbol{F}_{j} + \boldsymbol{R}_{M,2}^{T}\boldsymbol{M}_{j})}
\end{equation}
where \(\boldsymbol{R}_{M,1}^{T}\) and \(\boldsymbol{R}_{M,2}^{T}\) project \(\boldsymbol{F}_{j}\) and \(\boldsymbol{M}_{j}\) in the local coordinate system onto the global coordinate system, respectively. Let \(\boldsymbol{T}_{M}\) be the coordinate transformation matrix from the local coordinate system to the global coordinate system in Fig.\ref{figurelabel1}(c), which represents the local coordinates of the last node. Then
\begin{gather}
    \label{addequation17-19}
\boldsymbol{R}_{M,1}^{T} = \begin{bmatrix} \boldsymbol{0}_{3 \times (M-6)}  & \boldsymbol{T}_{M}^T & \boldsymbol{0}_{3 \times 3}
\end{bmatrix} \\
\boldsymbol{R}_{M,2}^{T} = \begin{bmatrix} \boldsymbol{0}_{3 \times (M-6)}  & \boldsymbol{0}_{3 \times 3}& \boldsymbol{T}_{M}^T
\end{bmatrix}
\end{gather}
Substituting (16) into (17) could get:
\begin{equation}
\label{eqn18}
\boldsymbol{F}_{a} = \sum_{j = 1}^{3}{(\boldsymbol{R}_{M,1}^{T} + \boldsymbol{R}_{M,2}^{T}\hat{\boldsymbol{d}_{j}})\boldsymbol{F}_{j}}
\end{equation}
where \(\hat{\boldsymbol{d}_{j}}\) is the skew-symmetric matrix of \(\boldsymbol{d}_{j}\).

\section{PROPOSED METHOD}\label{-proposed-method}

In this section, the mixed linear complementarity problem (MLCP) of FEM with contacts is introduced and an accelerated scheme is presented to solve the problem. Additionally, the parallelization on GPU of the update of the tangent stiffness matrix and the contact detection is described.

\subsection{Mixed Linear Complementarity Problem}\label{-mixed-linear-complementarity-problem}

To obtain the displacement of the continuum robot under the actuation force and contact force, it is necessary to solve by (\ref{eqn01}). Combining (\ref{eqn01}), (\ref{eqn02}), (\ref{eqn11}), (\ref{eqn12}), (\ref{eqn14}), (\ref{eqn15}) and (\ref{eqn18}) could get a Mixed Nonlinear Complementarity Problem (MNCP):
\begin{equation}
\label{eqn19}
\begin{cases}
\boldsymbol{F}\left( \boldsymbol{x}_{t} \right) + \boldsymbol{A}_{c}^{T}\boldsymbol{y}_{c} + \boldsymbol{A}_{fn}^{T}\boldsymbol{y}_{fn} - \boldsymbol{F}_{a} = \boldsymbol{0} \\
\boldsymbol{0} \leq \boldsymbol{y}_{c}\bot\boldsymbol{A}_{c}\Delta\boldsymbol{x}_{t} - \boldsymbol{b}_{c} \leq \boldsymbol{0} \\
\boldsymbol{y}_{fn}\bot\boldsymbol{A}_{fn}\Delta\boldsymbol{x}_{t} - \boldsymbol{b}_{fn} = \boldsymbol{0} \\
\boldsymbol{F}_{a} = \sum_{j = 1}^{3}{(\boldsymbol{R}_{M,1}^{T} + \boldsymbol{R}_{M,2}^{T}\hat{\boldsymbol{d}_{j}})\boldsymbol{F}_{j}}
\end{cases}
\end{equation}

Solving MNCP directly is complicated, but the nonlinear terms can be linearized such as $\boldsymbol{F}\left( \boldsymbol{x}_{t} \right)$.
Therefore the internal force at time \(t\) is linearized using the internal force and tangent stiffness matrix at time \(t - 1\):
\begin{equation}
\label{eqn20}
\boldsymbol{F}\left( \boldsymbol{x}_{t} \right) = \boldsymbol{F}\left( \boldsymbol{x}_{t - 1} \right) + \boldsymbol{K}\left( \boldsymbol{x}_{t - 1} \right)\Delta\boldsymbol{x}_{t}
\end{equation}
where \(\boldsymbol{K}\left( \boldsymbol{x}_{t - 1} \right)\) is the tangent stiffness matrix that depends on the actual position of the nodes at time \(t - 1\), which is updated using corotational formulation. The corotational formulation models the nonlinear deformation of flexible bodies by decomposing the motion of elements into rigid and pure deformed parts\cite{ref29},\cite{ref30}. This method allows to use of linear constitutive models for pure deformation, which means that the local stiffness matrix is constant and the parameters of different elements are relatively independent\cite{ref31},\cite{ref32}.

Substituting (\ref{eqn20}) into (\ref{eqn01}) could get:
\begin{equation}
\label{eqn21}
\boldsymbol{K}\left( \boldsymbol{x}_{t - 1} \right)\Delta\boldsymbol{x}_{t} + \boldsymbol{F}\left( \boldsymbol{x}_{t - 1} \right) - \boldsymbol{F}_{ext}\left( \boldsymbol{x}_{t} \right) = \boldsymbol{0}
\end{equation}
Then MNCP can be converted to a MLCP:
\begin{equation}
\label{eqn22}
\begin{cases}
\boldsymbol{K}\left( \boldsymbol{x}_{t - 1} \right)\Delta\boldsymbol{x}_{t} + \boldsymbol{F}\left( \boldsymbol{x}_{t - 1} \right) - \boldsymbol{F}_{ext} = \boldsymbol{0} \\
\boldsymbol{0} \leq \boldsymbol{y}_{c}\bot\boldsymbol{A}_{c}\Delta\boldsymbol{x}_{t} - \boldsymbol{b}_{c} \leq \boldsymbol{0} \\
\boldsymbol{y}_{fn}\bot\boldsymbol{A}_{fn}\Delta\boldsymbol{x}_{t} - \boldsymbol{b}_{fn} = \boldsymbol{0} \\
\boldsymbol{F}_{ext} = - \boldsymbol{A}_{c}^{T}\boldsymbol{y}_{c} - \boldsymbol{A}_{fn}^{T}\boldsymbol{y}_{fn} + \boldsymbol{F}_{a} \\
\boldsymbol{F}_{a} = \sum_{j = 1}^{3}{(\boldsymbol{R}_{M,1}^{T} + \boldsymbol{R}_{M,2}^{T}\hat{\boldsymbol{d}_{j}})\boldsymbol{F}_{j}}
\end{cases}
\end{equation}

\subsection{Solve the Problem}\label{-solve-the-problem}

Among the methods for directly solving Mixed Linear Complementarity Problems (MLCP), the commonly used Lemke method primarily focuses on constructing feasible solutions that meet the complementary conditions during iteration. However, the solution speed of this method is significantly impacted by the number of complementary constraints. Additionally, the convergence conditions of the Lemke algorithm are stringent, which limits its applicability. On the other hand, the GS iterative solver \cite{ref15} also utilizes complementary constraints for iteration, and its convergence rate is similarly affected by the number of constraints. Acc-FEM aims to address these issues by leveraging the semi-positive definite nature of the stiffness matrix to first transform the MLCP into a quadratic programming (QP) problem. By employing QP techniques, the iteration process is not always influenced by the number of constraints, thus mitigating the slow solution speed of MLCP in scenarios with multiple contacts while maintaining the accuracy of the algorithm.

In the MLCP, the force \(\boldsymbol{F}_{j}\) of \(j\)th cable is known and \(\Delta\boldsymbol{x}_{t}\) is the vector of nodal solution. To solve the problem, Karush--Kuhn--Tucker (KKT) conditions are used to transform the MLCP into a QP. Take \(\Delta\boldsymbol{x}_{t}\) as the unknown, take the first formula as the derivative of the QP objective function, and integrate and remove the constant term to get the objective function of QP. The second and third formulas are used as inequality constraints and equality constraints respectively. The last formula doesn\textquotesingle t change. That is the QP problem:
\begin{gather}
\min_{\Delta{x}_{t}}\frac{1}{2}\Delta{\boldsymbol{x}}_{t}^{T}\boldsymbol{K}\left( \boldsymbol{x}_{t - 1} \right)\Delta{\boldsymbol{x}}_{t} + {(\boldsymbol{F}\left( \boldsymbol{x}_{t - 1} \right) - \boldsymbol{F}_{a})}^{T}\Delta\boldsymbol{x}_{t} \nonumber\\
s.t.\ \boldsymbol{A}_{c}\Delta\boldsymbol{x}_{t} - \boldsymbol{b}_{c} \leq \boldsymbol{0} \nonumber\\
\boldsymbol{A}_{fn}\Delta\boldsymbol{x}_{t} - \boldsymbol{b}_{fn} = \boldsymbol{0} \nonumber\\
\boldsymbol{F}_{a} = \sum_{j = 1}^{3}{(\boldsymbol{R}_{M,1}^{T} + \boldsymbol{R}_{M,2}^{T}\hat{\boldsymbol{d}_{j}})\boldsymbol{F}_{j}} \label{eqn23}
\end{gather}

The solution of (\ref{eqn23}) is also the solution of (\ref{eqn22}). Under the elastic or hyperelastic assumption of deformations, the tangent stiffness matrix \(\boldsymbol{K}\left( \boldsymbol{x}_{t - 1} \right)\) is a symmetric positive semidefinite matrix, which allows most QP solvers to solve (\ref{eqn23}) and obtain a feasible solution. However, the large dimension of \(\boldsymbol{K}\left( \boldsymbol{x}_{t - 1} \right)\)significantly affects the computational speed of the simulation. Therefore, we adopt a sparse matrix representation for \(\boldsymbol{K}\left( \boldsymbol{x}_{t - 1} \right)\) to accelerate the computation. 

Equation (\ref{eqn23}) is a QP problem that can be solved by some existing quadratic programming packages, such as OSQP\cite{ref33}, Quadprog\cite{ref34}, Cvxopt\cite{ref35}, and so on. OSQP is a general-purpose solver for convex quadratic programs based on the alternating direction method of multipliers, which requires the solution of a quasi-definite linear system with the same coefficient matrix at almost every iteration\cite{ref33}. The reason for using OSQP is that the dimension of the quasi-definite linear system is the sum of the dimension of primal variables $\Delta\boldsymbol{x}_{t}$ and the number of constraints, meaning that the increase in the number of constraints has almost the same effect on the algorithm speed as the increase in the number of primal variables. Since the impact of increasing the number of constraints on the calculation speed is the same as the impact of increasing the primal variable dimension, when the primal variable dimension is already large, the increase in the number of constraints will have almost no effect, which can effectively solve the impact of multiple contacts.

\subsection{Parallelization on GPU}\label{-parallelization-on-GPU}

This section describes the parallelization of the update of MLCP and the contact detection used in Acc-FEM, which are implemented on GPU.

In the MLCP update, the main computational cost comes from the update of the tangent stiffness matrix, which needs to be updated after each deformation. Fortunately, in the finite element method, the tangent stiffness matrix is assembled from the local stiffness matrices of each element. Therefore, we only need to calculate the local stiffness matrices in parallel and assemble them into a tangent stiffness matrix.

Another crucial aspect of parallelization is contact detection. Acc-FEM must detect interactions between the robot and its environment. To achieve this, we match the nodes of the robot's finite element model with the triangles reconstructed from MRI images for collision detection. Each GPU thread performs these matchings, returns the collision position when contact is detected, and calculates the constraint as described in (3) in Section II. Both contact detection and constraint calculations are executed in parallel and subsequently integrated.

\subsection{Postprocessing for Contact Forces}\label{-postprocessing-for-contact-forces}

Due to the limitations of the contact detection method, the model has the risk of penetration, which may lead to the collapse of the algorithm. To reduce this risk, a method is proposed to limit step size, which is expressed as
\begin{equation}
\label{eqn25}
\Delta\boldsymbol{x}_{t}^{'} = \beta\Delta\boldsymbol{x}_{t}/\left\| \Delta\boldsymbol{x}_{t} \right\|
\end{equation}
where \(\beta\) is the step size related to \(\left\| \Delta\boldsymbol{x}_{t} \right\|\). The expression for \(\beta\) is as follows:
\begin{equation}
\label{eqn26}
\beta = \begin{cases}
\left\| \Delta\boldsymbol{x}_{t} \right\|,\ \ if\left\| \Delta\boldsymbol{x}_{t} \right\| \leq \beta_{0}\  \\
\beta_{0},\ \ \ \ \ \ \ \ if\left\| \Delta\boldsymbol{x}_{t} \right\| > \beta_{0}
\end{cases}
\end{equation}
where \(\beta_{0}\) is an artificial hyperparameter, which limits the step size. Correspondingly, \(\boldsymbol{y}_{c}\) and \(\boldsymbol{y}_{fp}\) decreases by the same factor:
\begin{gather}
\label{eqn27}
\boldsymbol{y}_{c}^{'} = \beta\boldsymbol{y}_{c}/\left\| \Delta\boldsymbol{x}_{t} \right\|
\boldsymbol{y}_{fn}^{'} = \beta\boldsymbol{y}_{fn}/\left\| \Delta\boldsymbol{x}_{t} \right\|
\end{gather}

Substituting (\ref{eqn25}), (\ref{eqn27}) and (\ref{eqn28}) into the first equation of (\ref{eqn22}) could get:
\begin{align}
\label{eqn28}
\boldsymbol{K}\left( \boldsymbol{x}_{t - 1} \right)\Delta\boldsymbol{x}_{t}^{'} & +  \boldsymbol{F}\left( \boldsymbol{x}_{t - 1} \right) + \boldsymbol{A}_{c}^{T}\boldsymbol{y}_{c}^{'} \nonumber \\
&+ \boldsymbol{A}_{fn}^{T}\boldsymbol{y}_{fn}^{'} - \boldsymbol{F}_{a}^{'} = \boldsymbol{0} 
\end{align}
Comparing the first equation of (\ref{eqn18}) and (\ref{eqn25}) could get:
\begin{equation}
\label{eqn29}
\left( \boldsymbol{F}_{a}^{'} - \boldsymbol{F}\left( \boldsymbol{x}_{t - 1} \right) \right)/\left( \boldsymbol{F}_{a} - \boldsymbol{F}\left( \boldsymbol{x}_{t - 1} \right) \right) = \ \beta/\left\| \Delta\boldsymbol{x}_{t} \right\|
\end{equation}

This postprocessing operation is equivalent to limiting the step size of displacement change and the step size of internal force change. When the sudden change of the actuation force of the continuum robot is too large, its displacement and internal force changes will be limited. Instead, more steps are used to calculate the change in displacement when a larger force is applied. Therefore, in the case of a burst, the calculation time of the algorithm is slightly increased but the frame rate does not change because of the intermediate variable result, and the stability of the algorithm is significantly improved.

\section{Experiments and Results}\label{-experiments-and-results}

This section presents several numerical experiments to evaluate the effectiveness of the Acc-FEM. Firstly, a comparative analysis was conducted to examine the computational speed of Acc-FEM and SOFA as the number of contacts increased. The second experiment compared Acc-FEM with a real robot to verify the accuracy of the algorithm. 

Python was used to implement the main algorithms, while the contact detection method was implemented using CUDA. The algorithms were run on a desktop equipped with an 11th-generation Intel (R) Core (TM) i9-11900 @ 2.50GHz processor and an NVIDIA GeForce RTX 3070 graphics card.

\subsection{Comparison Experiment with SOFA}\label{-comparison-experiment-with-sofa}

     \begin{figure}[t]
            \vspace{2mm}
		\centering
		\includegraphics[width=0.48\textwidth]{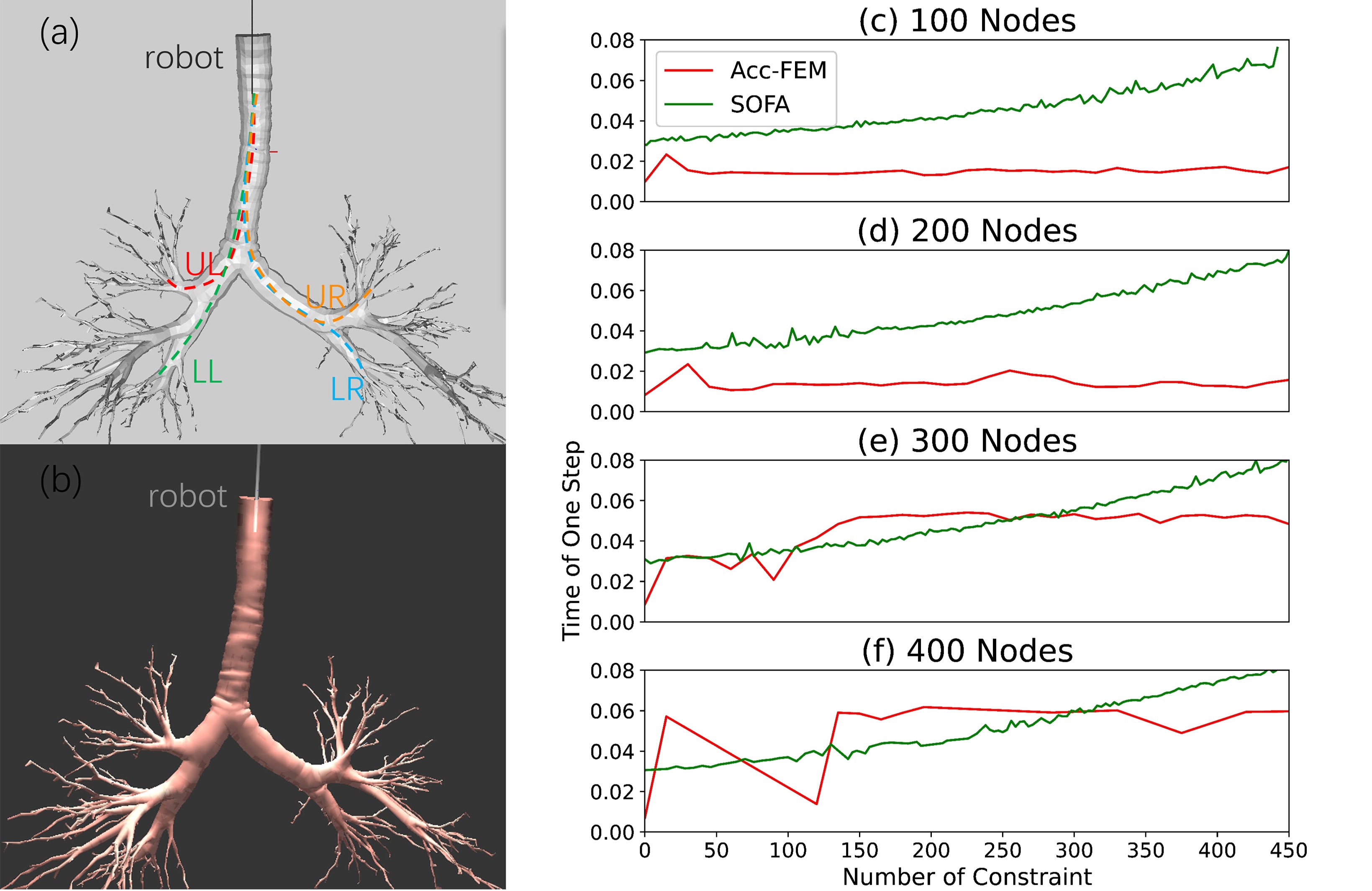}
		\caption{Comparison of Acc-FEM and SOFA. (a) Simulation setup for Acc-FEM. (b)    Simulation setup for SOFA. (c) Simulation results with 100 nodes. (d) Simulation results with 200 nodes. (e) Simulation results with 300 nodes. (f) Simulation results with 400 nodes.}
		\label{fig_sim5}
		\vspace{-2mm}
	\end{figure}

In the first experiment, Acc-FEM was compared with SOFA. For SOFA, we primarily employed the BeamAdapter \cite{ref36} module and the Gauss-Seidel (GS) solving method \cite{ref15}, while using default methods for other tasks such as collision detection. The BeamAdapter module is a standard tool in SOFA for multi-beam simulations, analogous to our discrete finite element method for beam elements. Acc-FEM was tested in the same scenario as SOFA, as shown in Fig.\ref{fig_sim5}, which depicted a simulated motion process of a continuum robot in the bronchi of the lungs. Fig.\ref{fig_sim5}(a) is our experimental setting, and Fig.\ref{fig_sim5}(b) is that of SOFA. The continuum surgical robot was divided into a different number of beam units with 100, 200, 300, or 400 nodes, which correspond to primal variables of 600, 1200, 1800, and 2400 dimensions, respectively. We then compared the calculation time of the two algorithms under a different number of contacts. To avoid the influence of other factors such as computer processing occupancy, we selected four different routes for the experiment, as shown in Fig.\ref{fig_sim5}(a), and took the average value of the calculation time. The results are presented in Fig.\ref{fig_sim5}, where the red curve and green curve represent the average calculation time of Acc-FEM and SOFA, respectively.

As depicted in Fig.\ref{fig_sim5}(c-f), SOFA demonstrates greater sensitivity to variations in contact quantity. As the number of contacts increases, the computation time per timestep of SOFA exhibits a more pronounced upward trend. Conversely, changes in the number of nodes exert less impact on SOFA. In contrast, Acc-FEM shows extended computation time as the number of nodes increases. However, it exhibits low sensitivity to variations in the dimensions of the number of contacts, resulting in smaller changes in computation time as the number of contacts increases.

This issue arises because the solver used in SOFA is the GS iterative method, which performs critical calculations within the contact space \cite{ref15}. In contrast, Acc-FEM converts the MLCP into a QP and employs the OSQP algorithm. This approach is less affected by an increase in constraints (equivalent to an increase in contacts), allowing Acc-FEM to maintain consistent calculation speed even as the number of contacts grows. In comparison, SOFA experiences a significant slowdown as the number of contacts increases. Additionally, Acc-FEM often achieves early convergence when the number of constraints is low, especially in scenarios with a large number of nodes and a small number of contacts, as demonstrated in Fig.\ref{fig_sim5}(e) and (f). This rapid convergence occurs because the algorithm handles fewer constraints more efficiently. In contrast, with a high number of constraints, the algorithm requires more iterations and thus more time to converge. Consequently, Fig.\ref{fig_sim5}(e) and (f) show that computation time decreases with fewer constraints, while with a larger number of constraints, computation time stabilizes, resulting in a leveling off of the curve.


     \begin{figure}[t]
            \vspace{2mm}
		\centering
		\includegraphics[width=0.48\textwidth]{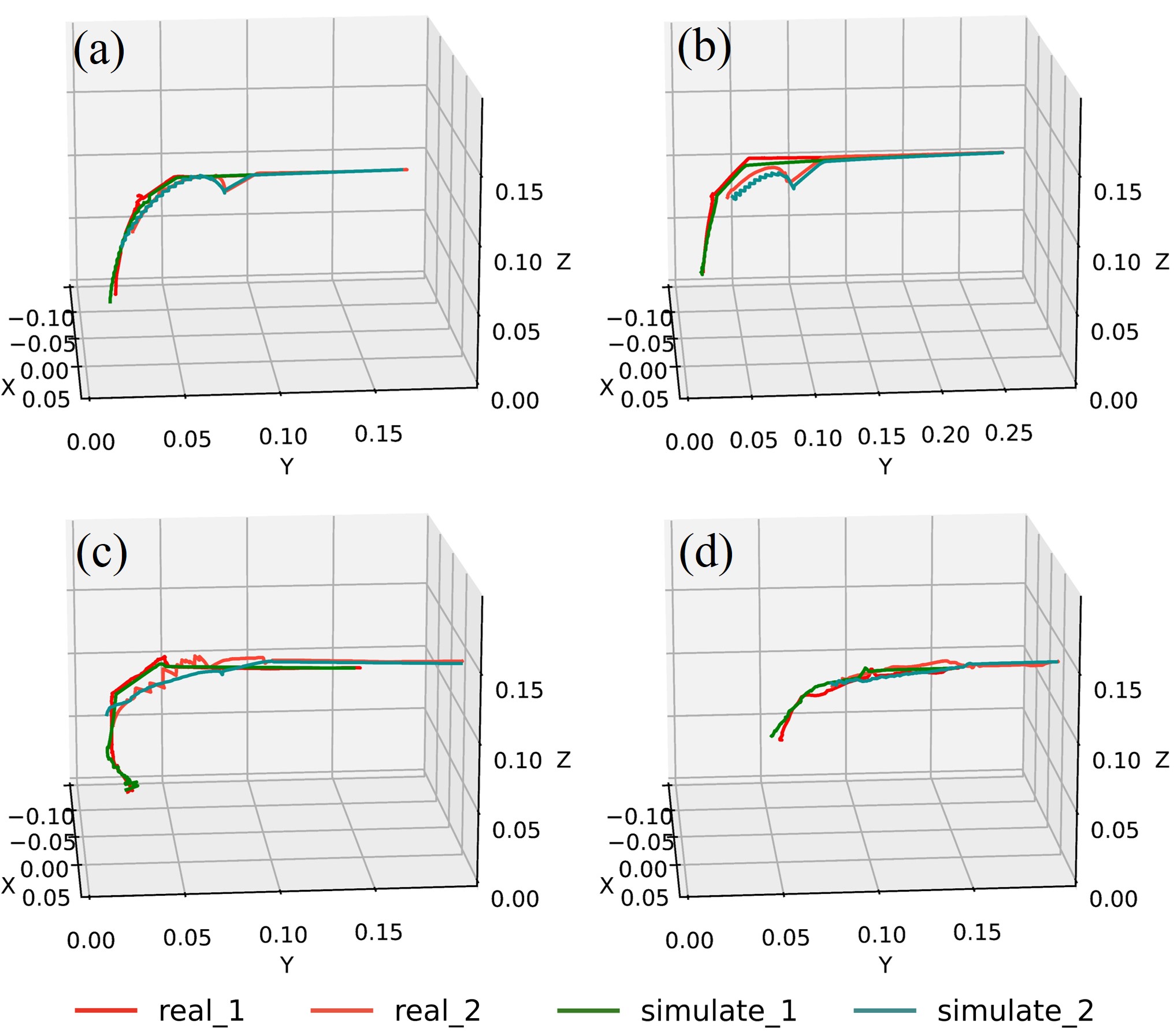}
		\caption{Experimental results of accuracy verification: (a) Tube with ‘T’-shaped model. (b) Tube with ‘T’-shaped model. (c) Robot with ‘T’-shaped model. (d) Robot with a bronchial model.}
		\label{fig_sim7}
		\vspace{-2mm}
	\end{figure}

    \begin{table*}[ht]
        \vspace{2mm}
        \centering
        \caption{Experimental Material Settings and Error Table}
        \begin{adjustbox}{max width=\textwidth}
        \begin{threeparttable}
        \small 
        \begin{tabular}{cccccccccc}
        \toprule
        \multirow{2}{*}{Experiment No.} & \multicolumn{4}{c}{Material Settings} & \multicolumn{4}{c}{Error} \\
        \cmidrule(lr){2-5} \cmidrule(lr){6-9}
        & Young's modulus/$MPa$ & Cross-sectional area/$m^2$ & Moment of inertia/$m^4$ & Location of sensors/$m$ & Sensor 1 Max Error/$m$ & Sensor 1 Avg Error/$m$ & Sensor 2 Max Error/$m$ & Sensor 2 Avg Error/$m$ \\
        \midrule
        a & 510 & 5.551e-6 & 5.041e-12 & 0.0206 / 0.06781\tnote{*} & 0.0058 & 0.0013 & 0.0049 & 0.002 \\
        b & 307 & 9.03e-6 & 19.233e-12 & 0.018 / 0.0754\tnote{*} & 0.0061 & 0.0029 & 0.0058 & 0.0022 \\
        c & 510 & 5.551e-6 & 5.041e-12 & 0.0088 / 0.0728\tnote{*} & 0.0066 & 0.0027 & 0.0124 & 0.0032 \\
        d & 510 & 5.551e-6 & 5.041e-12 & 0.0088 / 0.0728\tnote{*} & 0.0065 & 0.0034 & 0.0076 & 0.0037 \\
        \bottomrule
        \end{tabular}
        \begin{tablenotes}
        \item[*] {The distance between the sensor and the end. The two numbers represent sensor 1 and sensor 2 respectively.}
        \end{tablenotes}
        \end{threeparttable}
        \end{adjustbox}
        \vspace{-4mm} 
        \label{tab1}
    \end{table*}

\subsection{Accuracy Verification In Real Robot} \label{-accuracy-verification-in-real-robot}

In the second experiment, we compared the performance of Acc-FEM against that of a real continuum robot operating in the same environment. The experimental setup for the real robot is depicted in Fig.\ref{figurelabel1}(a). The flexible robot was mounted on a robotic arm, which was used to move the robot and perform operations such as insertion and extraction. We had positioned two electromagnetic tracking sensors (10001742, NDI Aurora) at the end of the robot to accurately capture its real-time position. During the experiment, we performed insertion operations with the real robot in a rigid environment, recording both the sensor positions and the input operations. Subsequently, we replicated the same operations in the simulator and recorded the positions of the points on the simulated robot corresponding to the electromagnetic tracking sensors. The material parameters for the robot and rubber tube utilized in the experiments are detailed in Table \ref{tab1}. The lengths of all the tubes or robot were 470mm, and the length of the flexible section at the end of the robot was 70mm. To evaluate the accuracy of the algorithm, we conducted four experiments, labeled (a), (b), (c), and (d). The first two experiments (a) and (b) employed various cable-free rubber tubes, while the last two experiments (c) and (d) used our robot.

The experimental results are shown in Fig. \ref{fig_sim7}(a-d). The red and light red lines indicate the positions of the two sensors on the real robot, while the blue and green lines show the corresponding positions in the algorithm simulation. For the cable-free rubber tubes, accuracy is relatively high, as seen in Fig. \ref{fig_sim7}(a) and (b), with nearly identical sensor curves indicating precise results from Acc-FEM. However, substantial error is observed in Fig. \ref{fig_sim7}(c) and (d). The simulation in Fig. \ref{fig_sim7}(a) and (b) aligns closely with real sensor positions, confirming Acc-FEM’s accuracy despite variations in cable control. Fig. \ref{fig_sim7}(c) shows a jagged curve in the middle stage due to significant friction between the robot and the ‘T’-shaped hollow plastic model, which wasn't simulated. Nonetheless, the alignment of the zigzag pattern at the end and middle of sensor 1 demonstrates Acc-FEM's ability to handle robot control and large deformations. In contrast, Experiment (d) shows a more straightforward path due to limited robot movement within the bronchial model. Overall, Acc-FEM effectively simulates the robot's behavior under various contact and control conditions.

Table \ref{tab1} lists the maximum and average errors for the four experiments. The maximum error of 0.0124 meters ($2.6\%$ of the total length) occurs on sensor 2 in Experiment (c), where a zigzag path contributes to the error. Experiments (a) and (b) have relatively small errors due to the absence of cable effects. Experiment (d) shows the highest average error (0.0034m and 0.0037m, or $0.72\%$ and $0.79\%$ of the total length), influenced by both cable effects and the complex environment. Overall, the errors are small, demonstrating the accuracy of Acc-FEM.

\section{Conclusion}\label{-conclusion}

In this paper, we propose the Acc-FEM, an accelerated FEM of continuum robots. We introduce the cable-driven continuum robot and provide the finite element modeling method and quasi-static physical model for multi-contact scenarios. Building upon this, we propose a solver scheme and a parallelization method. We design several experiments to validate Acc-FEM and make comparisons with other schemes, such as SOFA. The experiments prove the effectiveness and superiority of Acc-FEM, which significantly improves the real-time performance of the finite element model. However, in this article, we did not consider the issue of environmental deformation, which will be one of our future research focuses.

In our future work, we aim to further enhance the modeling of the continuum robot by transitioning from a quasi-static model to a dynamic model. 






\bibliographystyle{ieeetr}
\bibliography{ref}

\end{document}